\documentclass{amia}
\usepackage{graphicx}
\usepackage[labelfont=bf]{caption}
\usepackage[superscript,nomove]{cite}
\usepackage{color}
\usepackage[sort,comma,numbers,super]{natbib}
\usepackage[dvipsnames]{xcolor}
\usepackage{soul}
\usepackage{hyperref}
\usepackage[rightcaption]{sidecap}
\usepackage[dvipsnames]{xcolor}
\usepackage{wrapfig}

\newcommand\EatDot[1]{}

\definecolor{green_c}{HTML}{2CA02C}
\definecolor{blue_c}{HTML}{1F77B4}
\definecolor{purple_c}{HTML}{9467BD}

\begin{document}

\title{RadLex Normalization in Radiology Reports}

\author{Surabhi Datta, MS, Jordan Godfrey-Stovall, BS, Kirk Roberts, PhD}

\institutes{
    School of Biomedical Informatics \\
    The University of Texas Health Science Center at Houston \\
    Houston, TX \\
}

\maketitle

\noindent{\bf Abstract}

\textit{Radiology reports have been widely used for extraction of various clinically significant information about patients’ imaging studies.  However, limited research has focused on standardizing the entities to a common radiology-specific vocabulary. Further, no study to date has attempted to leverage RadLex for standardization. In this paper, we aim to normalize a diverse set of radiological entities to RadLex terms.  We manually construct a normalization corpus by annotating entities from three types of reports. This contains 1706 entity mentions.  We propose two deep learning-based NLP methods based on a pre-trained language model (BERT) for automatic normalization. First, we employ BM25 to retrieve candidate concepts for the BERT-based models (re-ranker and span detector) to predict the normalized concept.  The results are promising, with the best accuracy (78.44\%) obtained by the span detector.  Additionally, we discuss the challenges involved in corpus construction and propose new RadLex terms.}

\section*{Introduction}
Radiology reports contain a wide range of entities describing the interpretations of the corresponding images. Prior research \cite{steinkamp2019CompleteStructuredInformation, bozkurt2019AutomatedDetectionMeasurements,fu2020DevelopmentClinicalConcept,datta2020RadSpatialNetFramebasedResourcea} has focused on developing methods to identify clinically-significant information from these reports. They emphasize using this extracted information in a variety of downstream clinical applications including automated tracking of abnormal radiographic findings (e.g., lesions), summarization, and cohort selection for epidemiological research. However, to enable the use of the extracted entities in the process of developing the automated systems across multi-institutional reports, the entities need to be mapped to concepts in a standardized vocabulary of radiology terms. There has been limited efforts in this direction, and therefore we aim to normalize the different entities or information types to RadLex \cite{langlotz2006radlex} concepts to facilitate improved consistency in the structured representations of important radiological entities.

\begin{SCfigure}[][b]
\includegraphics[width=0.5\textwidth]{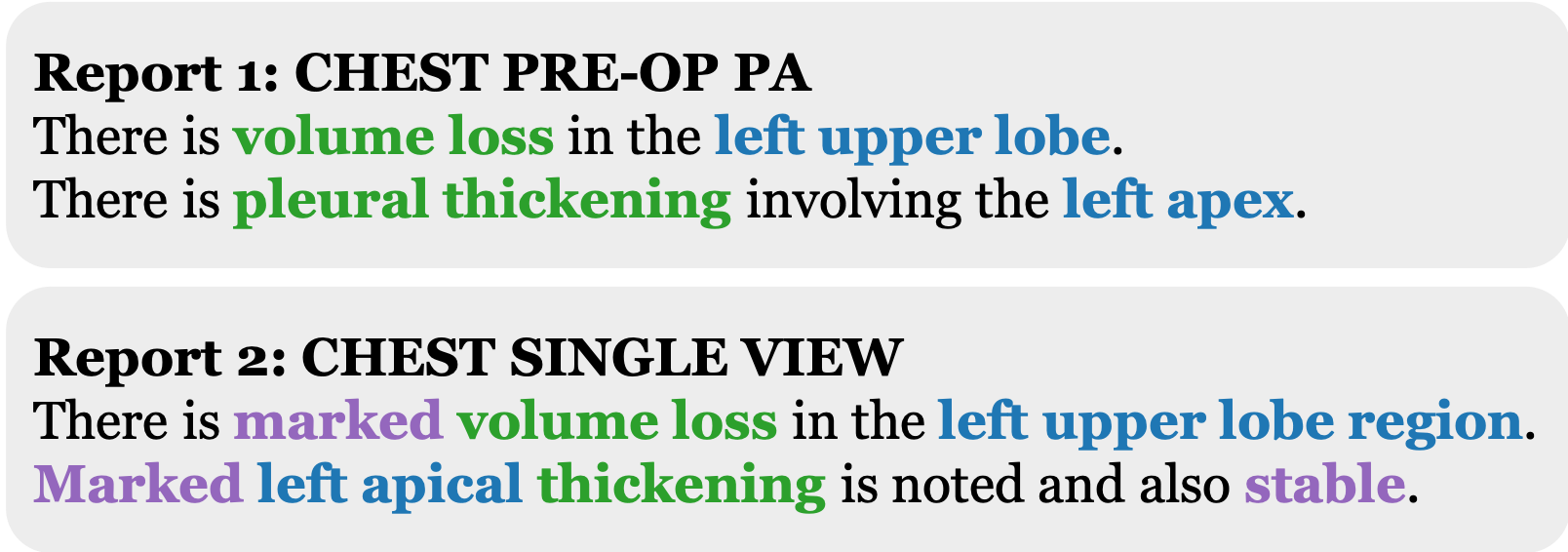}
% \vspace{-0.2in}
\centering
\caption{Partial section of two radiology reports. Findings are shown in \textcolor{green_c}{green}, anatomical locations are in \textcolor{blue_c}{blue}, and the descriptor terms in \textcolor{purple_c}{purple}. The RadLex concepts corresponding to the two anatomical locations are upper lobe of left lung (RID1327) and left lung (RID1326) + apex (RID5946)}.
\label{fig:report_examples}
\end{SCfigure}

Radiologists use different phrases to express the same concept in a report. Normalization is the process of mapping these phrase spans in text to standard concepts in a vocabulary. For instance, \textit{Right base} and \textit{Right lower lung zone} are different forms of describing the same anatomical entity. Similarly, \textit{Intramural or free air} is used by radiologists to indicate the clinical finding - \textit{Pneumatosis intestinalis}, and \textit{Central line}s are often used to denote the presence of \textit{Central venous catheter}s in chest X-ray reports. Many natural language processing (NLP)-based clinical application systems rely on developing inference rules. Such inferences are often performed on the entities that are extracted from clinical text by entity recognition systems. Consider the example shown in Figure \ref{fig:report_examples} to illustrate the usability of normalizing radiological entities in the report text for automated abnormality tracking systems.

The sentences in Figure \ref{fig:report_examples} appear in two different reports of a patient. \textit{Volume loss} and \textit{Pleural thickening} are the main finding-related entities while \textit{left upper lobe} and \textit{left apex} are the anatomy-related entities in the first report. The second report also contains the same findings except that the anatomical phrases are changed to \textit{left upper lobe region} and \textit{left apical} and there are additional clinically-relevant status descriptors such as \textit{marked} and \textit{stable}. Utilizing the extracted contextual information (e.g., anatomical phrases here) in the automated tracking of \textit{volume loss} and \textit{thickening} would require establishing a standardized way to represent the extracted anatomical entities (i.e., mapping both \textit{left upper lobe} and \textit{left upper lobe region} to RadLex concept \textbf{\textit{upper lobe of left lung (RID1327)}}). To our knowledge, only one study \cite{tahmasebi2019AutomaticNormalizationAnatomical} so far has worked on normalizing the anatomical terms in the reports to SNOMED CT ontology. In this paper, we attempt to broaden the scope of entity types and consider all those that can act as contextual information in various potential clinical use cases. Moreover, since RadLex \cite{langlotz2006radlex} is a publicly available radiology lexicon (containing 46,657 concepts) specifically developed for standardizing the language used in reporting imaging results, we utilize RadLex for mapping the various entity mentions in the reports. This will cover entities such as common modifiers and uncertainty phrases often encountered in radiology report text and may not be present in other ontologies such as SNOMED CT.

In this work, we create a manually-annotated corpus \textcolor{black}{of radiology reports} covering three different imaging modalities from the MIMIC-III clinical note corpus \cite{Johnson2016mimicIII} for concept normalization. We describe the detailed process of annotating different radiological entities. The entity types include clinical finding, imaging observation, anatomical phrase, medical device, different descriptor terms (e.g., status, temporal, certainty), imaging procedure, imaging modality, process, and property-related information. This is the first study focusing on the normalization of diverse radiological entities to a standard radiology-specific lexicon -- RadLex. Our annotated corpus consists of $1706$ entity mentions and $449$ distinct RadLex concepts, with the frequent entity types being \textcolor{black}{RadLex descriptors} and \textcolor{black}{anatomical entities}. There are a total of $151$ entities in the dataset that are unlinkable (i.e., cannot be mapped to any existing RadLex code). We also refer to these unlinkable entities as RID-less when discussed in context to normalization annotations.
Instructions to access the corpus are available at \url{https://github.com/krobertslab/datasets}.
% KIRK: are we going to include the URL here?  If this is MIMIC data, then we likely need to upload the corpus to PhysioNet instead of sharing on github.

\textcolor{black}{For normalizing the radiological entities, we first expand each entity mention using synonym information in RadLex. We also expand the abbreviated entity mentions. We apply the BM25 information retrieval technique \cite{robertson1996OkapiTREC3} to generate candidate RadLex concepts for each of the expanded entity mentions. Then we propose two methods based on a pre-trained transformer-based language model, BERT \cite{devlin2019BERTPretrainingDeep}, for selecting the final normalized concept from the set of candidate concepts. The first method incorporates target lexicon knowledge from RadLex to obtain the normalized concept by re-ranking the candidates generated by BM25. The second method, where BERT is utilized analogous to the setting of a question answering task, selects the normalized concept given a radiological entity mention and the list of BM25 candidates for that mention. Since BM25 can retrieve any concept in the RadLex lexicon as a candidate concept, this method addresses the limitations of supervised classification techniques where the normalized concept lies within the scope of concepts in the training corpus. Although the main focus of our work is RadLex normalization, we also examine the performance of BERT in automatically detecting the entity spans from the reports.}

Additionally in this paper, we address some of the challenges that emerged while creating the annotated corpus for radiology concept normalization. We also propose a list of new RadLex terms particularly focusing on chest X-rays, Brain MRIs, and babygram-related imaging studies which can serve as a resource to the research community for expanding RadLex in the future.

\section*{Related Work}
\section{Annotated corpora for Medical Concept Normalization}

The NCBI disease corpus, consisting of 793 PubMed abstracts, was annotated for disease name normalization \cite{dogan2014NCBIDiseaseCorpus}. There exists an annotated dataset of narrative clinical reports for the normalization task of disorders as part of ShARe/CLEF eHealth 2013 challenge\footnote{\url{https://sites.google.com/site/shareclefehealth/}}. Disorder normalization corpora constructed from MIMIC clinical notes are also available through SemEval-2014 Task 7\footnote{\url{http://alt.qcri.org/semeval2014/task7/}} and SemEval-2015 Task 14\footnote{\url{http://alt.qcri.org/semeval2015/task14/}}. A few studies created medical concept normalization corpora for mapping user generated text on social media to standard vocabularies like SNOMED. CADEC consists of annotated concepts from 1253 social media posts taken from AskaPatient\footnote{\url{https://www.askapatient.com/}} associated with adverse drug events (ADEs) of patients \cite{karimi2015CadecCorpusAdverse}. PsyTAR, also constructed from AskaPatient, contains 887 patient posts annotated with ADEs related to psychiatric medications \cite{zolnoori2019SystematicApproachDeveloping}. Sarker et al.\cite{sarker2018DataSystemsMedicationrelated} developed an annotated corpus for normalizing expressions denoting adverse drug reactions (ADRs) from Twitter to MedDRA \cite{brown1999MedicalDictionaryRegulatory} (Medical Dictionary for Regulatory Activities) Preferred Terms (PTs), which was released in the shared task -- Social Media Mining for Health (SMM4H). \textcolor{black}{Roberts et al.\cite{roberts2017OverviewTAC2017} also released an annotated dataset of 200 drug labels for TAC2017 where the ADR expressions in the labels were mapped to MedDRA Lower Level Terms and PTs.} A recent study has also released an annotated corpus of 100 discharge summaries in a 2019 shared task covering entities corresponding to medical problems, treatments, and tests \cite{luo2019MCNComprehensiveCorpus}. Previous works have mapped the concept mentions to ontologies such as the Unified Medical Language System (UMLS) \cite{bodenreider2004UnifiedMedicalLanguage}, SNOMED, RxNorm, and MedDRA. Thus, we note that no work has focused on constructing normalization corpus from radiology reports and mapping the important radiological entity spans to RadLex codes.

\section{Methodologies employed for Medical Concept Normalization using Deep Learning}

Some of the first deep learning approaches for concept normalization in the medical domain were based on convolutional neural networks (CNNs) and recurrent neural networks (RNNs) where a user phrase was converted to a semantic vector representation and eventually a softmax classifier was used to assign a standard medical concept to that phrase \cite{limsopatham2016NormalisingMedicalConcepts,tutubalina2018MedicalConceptNormalization,han2017TeamUKNLPDetecting}. Tutubalina et al. \cite{tutubalina2018MedicalConceptNormalization}, however, incorporated additional semantic similarity features by leveraging prior domain knowledge (UMLS) to further enrich the phrase representations. A recent study by Miftahutdinov et al. \cite{miftahutdinov2019DeepNeuralModels} used contextualized word representations such as BERT and ELMo \cite{peters2018DeepContextualizedWord} for normalizing user generated phrases and achieved state-of-the-art performance on three benchmark normalization datasets -- CADEC, PsyTAR, and SMM4H 2017. All these papers worked on user-generated text of social media posts and formulated normalization as a multi-class classification task. Another work \cite{luo2019HybridNormalizationMethod} has proposed a hybrid system by combining exact match, edit-distance, and deep learning methods for normalizing entities in the ShARe/CLEF 2013 challenge dataset. Their model architecture additionally integrated contextual information of an entity mention (left and right context words) and predicted the UMLS code using a softmax classification layer. 

However, Ji et al. \cite{ji2019BERTbasedRankingBiomedical} have used BERT as a ranking model in a normalization task. They ranked the candidate concepts after generating them using the BM25 \cite{robertson1996OkapiTREC3} information retrieval method. Their BERT-based ranker outperformed the previous best results on ShARe/CLEF, NCBI, and TAC2017ADR normalization datasets. Moreover, BERT-based re-ranking has been shown to perform well on other information retrieval tasks such as passage retrieval \cite{nogueira2019PassageRerankingBERT}. \textcolor{black}{Inspired by these, we propose to utilize a BERT-based re-ranker along with incorporating domain knowledge for radiology entity normalization. Additionally, we propose to formulate the normalization task by adopting BERT in a configuration that has been predominantly used for question answering task \cite{devlin2019BERTPretrainingDeep}.}

To the best of our knowledge, due to the lack of annotated radiology corpus, no study so far has applied supervised learning techniques for radiology concept normalization. In view of the promising results of applying deep learning methods for normalization, we aim to create a comprehensive annotated radiology normalization dataset in this work and apply deep learning-based method on the dataset. A previous study \cite{tahmasebi2019AutomaticNormalizationAnatomical} has utilized an unsupervised semantic learning approach to normalize the anatomical phrases in the radiology reports to SNOMED CT anatomical concepts. However, their work was limited to normalizing only the anatomical terms and did not cover other commonly observed clinically-significant information such as clinical findings and modifier terms.

\section*{Materials and Methods}
\setcounter{section}{0}
\section{Dataset}
\begin{figure*}[t]
\includegraphics[width=0.6\textwidth]{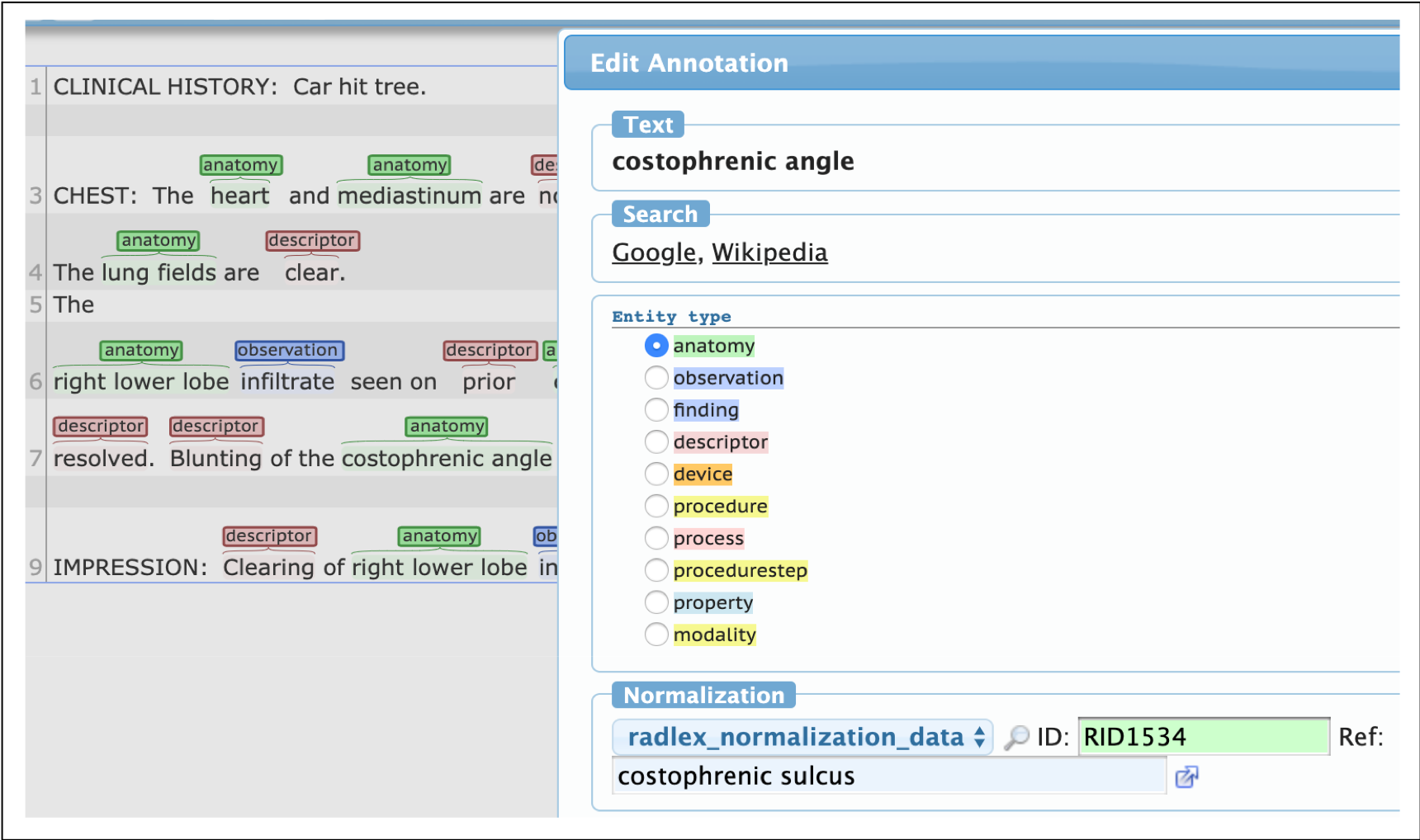}
% \vspace{-0.2in}
\centering
\caption{Example annotation to normalize ``\textit{costophrenic angle}'' to RadLex term ``\textit{costophrenic sulcus}'' corresponding to RadLex ID RID1534 in a sample report using BRAT 1.3.}
\label{fig:annotation}
\end{figure*}

We selected a subset of 50 radiology reports from MIMIC-III clinical database \cite{Johnson2016mimicIII}. This consists of \textcolor{black}{17} chest X-ray reports, \textcolor{black}{16} Brain Magnetic Resonance Imaging (MRI) reports, and \textcolor{black}{17} babygram-related reports. This set of reports covers some common imaging techniques and a wide range of anatomical locations (as often babygrams contain descriptions of multiple body organs). We used the BRAT annotation tool \cite{stenetorp2012BratWebbasedTool} for annotating the radiological entities with their corresponding RadLex IDs as shown in Figure \ref{fig:annotation}.

% \pagebreak

\section{Annotation}
\subsection{Identifying Entity Spans}

The first annotation task is to identify the entity mention in the report sentences whose type falls under one of the following broad RadLex classes:
\begin{enumerate}
\vspace{-0.2in}
  \item \textsc{\textbf{Clinical Finding}} - Refers to pathophysiologic finding, and symptoms (e.g., \textit{heart failure})
  \vspace{-0.1in}
  \item \textsc{\textbf{Imaging Observation}} - Image-specific features as interpreted by radiologists (e.g., \textit{infiltrate})
  \vspace{-0.1in}
  \item \textsc{\textbf{Anatomical Entity}} - Refers to a body location (e.g., \textit{apex of lung})
  \vspace{-0.1in}
  \item \textsc{\textbf{Medical Device}} - Refers to a medical object (e.g., \textit{endotracheal tube})
  \vspace{-0.1in}
  \item \textsc{\textbf{RadLex Descriptor}} - Any modifier (usually adjectives) used to describe other entities like clinical finding (e.g., status descriptor - \textit{stable}, composition descriptor - \textit{osseous}, certainty descriptor - \textit{no}, etc.)
%   size descriptor - \textit{dilated}, etc.)
  \vspace{-0.1in}
  \item \textsc{\textbf{Procedure}} - This includes different procedures such as imaging procedures, follow-up procedures, and treatment. (e.g., \textit{catheter removal})
  \vspace{-0.1in}
  \item \textsc{\textbf{Procedure Step}} - Includes any step in image processing (e.g., \textit{multiplanar reformat})
  \vspace{-0.1in}
  \item \textsc{\textbf{Process}} - Usually refers to treatment planning, change etc. (e.g., \textit{motion})
  \vspace{-0.1in}
  \item \textsc{\textbf{Imaging Modality}} - Form of imaging that depends on how the image is produced (e.g., \textit{magnetic resonance imaging})
  \vspace{-0.1in}
  \item \textsc{\textbf{Property}} - Modifier terms (usually noun phrases) associated with entities (e.g., \textit{patient rotation 
  position})
\end{enumerate}

\subsection{Instructions for Assigning RadLex Codes}

The next step involves assigning a single RadLex ID to each of the identified entity mentions. Note that we have mapped each entity to only one RadLex ID. For instance, the anatomical entity ``\textit{Midline structure}'' is mapped to the RadLex concept ``\textit{Septum pellucidum}'' with RadLex ID RID6525. While assigning the RadLex ID, the following instructions were given to the annotators:
\begin{enumerate}
\vspace{-0.15in}
  \item Search for the exact entity span in RadLex
  \vspace{-0.1in}
  \item If not found using 1, search whether it appears in RadLex with different a variation such as with words rearranged in a different order (e.g., assigning RadLex concept \textit{apex of lung} for the entity mention \textit{lung apex})
  \vspace{-0.1in}
  \item If not found using 1 and 2, search whether it appears as a synonym or in the decription of another RadLex concept (e.g., \textit{Costophrenic sulcus} is present as a synonym of the RadLex concept \textit{Costophrenic angle})
  \vspace{-0.1in}
  \item If not found using the above, refer the web to look for the most semantically similar concept in RadLex (e.g., \textit{Chest tube} is mapped to the RadLex concept \textit{Thoracostomy tube} following this guideline)
  \vspace{-0.1in}
  \item If an entity cannot be mapped to any RadLex concept, it is assigned a label ``RID-less''
\end{enumerate}
\vspace{-0.1in}
Moreover, while annotating entities following the above instructions, the following points are taken into consideration:
\vspace{-0.25in}
\begin{itemize}
  \item \textbf{Taking context into account} - Annotation of some entities may vary based on the context of the sentence or the anatomical entity associated with the imaging modality. For example, \textit{Microangiopathic changes} refers to a disease/condition affecting small blood vessels. So depending on the anatomical entity associated with the imaging modality, the mapping would vary. When the imaging results are related to heart, \textit{Microangiopathic changes} would be mapped to \textit{Microvascular ischemia} in RadLex which is listed under \textit{Cardiovascular disorder}, whereas for brain-related imaging results \textit{Microangiopathic changes} will not be assigned any RadLex code.
  \item \textbf{Splitting entity spans to subspans} - Many times an entity mention cannot be mapped to a specific RadLex concept and annotators tend to use multiple RadLex concepts in different combinations to annotate that entity. For example, \textit{Right middle lobe} is not directly normalizable to a RadLex concept and may be mapped to concepts like \textit{lobe}, \textit{middle lobe of lung}, or \textit{right}. In order to resolve ambiguity in the annotation process, each entity mention is split such that the split with the largest subspan can be mapped to a RadLex code and all the other smaller subspans are also mapped to their corresponding RadLex codes. Thus, ``\textbf{Right middle lobe}'' will be split as ``\textbf{middle lobe}'' (largest RadLex mappable subspan) +  ``\textbf{right}'' and not as ``\textbf{right}''+ ``\textbf{middle}'' + ``\textbf{lobe}''. Further, ``middle lobe'' will be normalized to ``middle lobe of lung'' (RID1310) and ``right'' to ``right'' (RID5825).
  
  However, there may arise cases when multiple possible variations of largest mappable subspans can be generated. For example,  ``Right lung apex'' results in two valid (RadLex mappable) splits, one with ``Right lung'' + ``apex'' and the other with ``Right'' + ``lung apex''. This can be resolved during reconciliation phase by incorporating domain knowledge (e.g., knowledge on human anatomy) and further verification by a physician. This may favor the first option -- ``Right lung'' + ``apex'' as this is more close to describing the apex of right lung.
\end{itemize}

% \pagebreak

\begin{wraptable}{r}{0.5\linewidth}
\footnotesize
\vspace{-0.1in}
\caption{Descriptive statistics of the annotated corpus.}
\vspace{-0.1in}
\centering
\begin{tabular}{l|c}
  \hline
 \textbf{Item} & \textbf{Frequency} \\ 
  \hline
 \textsc{Clinical Finding} & $282$ \\
 \textsc{Imaging Observation} & $77$ \\
 \textsc{Anatomical Entity} & $384$ \\
 \textsc{Medical Device} & \textcolor{black}{$102$} \\
 \textsc{RadLex Descriptor} & \textcolor{black}{$651$} \\
 \textsc{Procedure-Related} & \textcolor{black}{$46$} \\
 \textsc{Process} & \textcolor{black}{$28$} \\
 \textsc{Imaging Modality} & \textcolor{black}{$51$} \\
 \textsc{Property} & \textcolor{black}{$85$} \\
  \hline
  Total entity mentions & \textcolor{black}{$1706$} \\
  \hline
  Unlinkable mentions & \textcolor{black}{$151$} \\
  \hline
  
\end{tabular}
\label{table:descriptive_stats}
%\end{table}
\end{wraptable}

Every report was double-annotated and reconciled with the clinical knowledge verified by a physician when required.
The F1 agreement between the two annotators (S.D. and J.S.) in annotating the spans of radiological entity mentions is $0.60$. We considered an exact match in the entity spans for calculating the F1 score.
The normalization agreement (accuracy) between the annotators on the reconciled version of the entity mentions is $76.7$\%.
Basic statistics of our annotated corpus are shown in Table~\ref{table:descriptive_stats}.

\section{Methods}

\subsection{Entity span detection}
\label{entity_span_detection}
We formulate this as a sequence labeling task where each word that is part of any radiological entity of interest is tagged using Beginning and Inside tags whereas a word that is not a part of an entity is tagged as Outside.
Each sentence in the reports is WordPiece-tokenized. This tokenized sentence is represented as [[CLS] sentence [SEP]] following the original paper \cite{devlin2019BERTPretrainingDeep} and then fed into the BERT model.

\vspace{-0.1in}

\subsection{Normalization methods}
\begin{figure*}[t]
\includegraphics[width=1\textwidth]{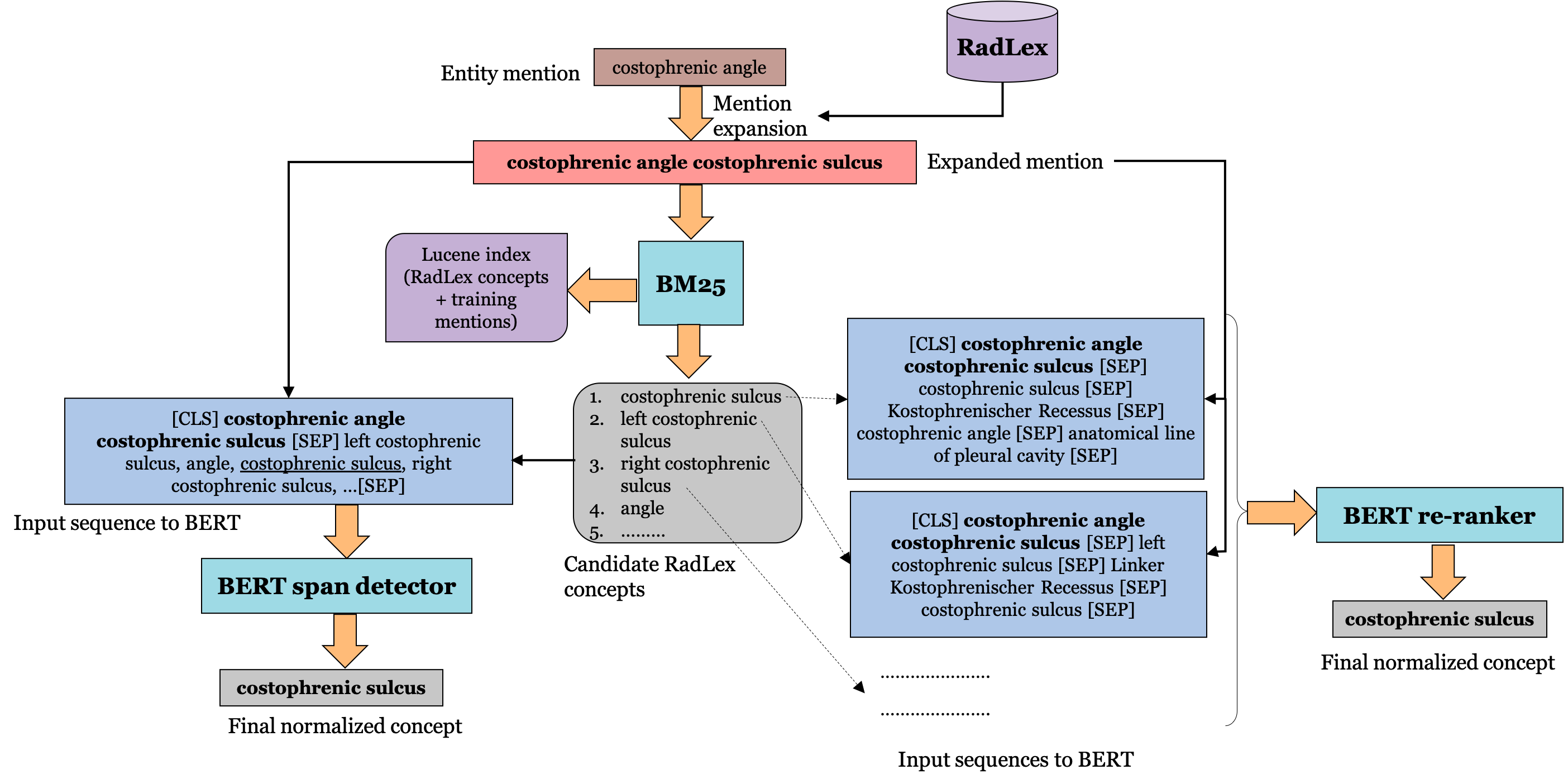}
\vspace{-0.2in}
\centering
\caption{Overview of the normalization process using the proposed methods (demonstrated for the entity mention-``\textit{costophrenic angle}'').}
\label{fig:pipeline}
\end{figure*}

The following subsections contain the descriptions of the three methods used for RadLex normalization. The overall framework of the normalization methods is illustrated in Figure~\ref{fig:pipeline}.

\vspace{-0.1in}
\subsubsection{BM25}
We index all RadLex concepts (a total of \textcolor{black}{$46,657$} Preferred Names in RadLex) as well as the entity mentions present in the training sets of our annotated corpus using Anserini \cite{yang2018AnseriniReproducibleRanking}.
We then use BM25 to retrieve and initially rank a set of $n$ candidates for each entity mention.
(In our experiments, we use $n=10$.)
We set the values of BM25 parameters, $b$ and 
$k1$, as 0.75 and 1.2, respectively.
In order to maximize the recall of BM25 in the candidate generation phase, each entity mention is transformed using the following two expansion techniques:
\vspace{-0.1in}
\begin{enumerate}
    \item Using \textbf{\textit{Synonyms}} in RadLex - If the entity mention ($m$) appears as a Synonym of a RadLex concept ($r_{c}$), the original mention is expanded using the Synonym.
    For example, ``encephalopathy'' is not present in RadLex but appears as a Synonym of the RadLex concept ``disorder of brain'' (RID5055).
    Thus, the mention ``encephalopathy'' is expanded to ``encephalopathy disorder of brain''.
    \vspace{-0.1in}
    \item \textbf{\textit{Abbreviation}} expansion - Often, some common medical devices and clinical findings are abbreviated in the reports.
    We expand these mentions leveraging the medical abbreviations and acronyms of radiopaedia\footnote{\url{https://radiopaedia.org/articles/medical-abbreviations-and-acronyms-a?lang=us}}.
    For instance, ``NGT'' is expanded to ``nasogastric tube'' and ``NPH'' is expanded to ``normal pressure hydrocephalus''.
\end{enumerate}
\vspace{-0.2in}

\subsubsection{BERT as re-ranker}
We use the set of candidate concepts obtained from BM25 for each entity mention to train the BERT\textsubscript{BASE} model for re-ranking these candidates.
The highest ranked candidate predicted by BERT is chosen as the final normalized RadLex concept for a given radiology entity mention. 
The model is trained as a binary classification task where for each candidate concept ($c_{i}$) and the mention ($m$) pair, the label is assigned as 1 when the candidate is the actual annotated normalized concept for that mention. 
For each such pair of candidate concept and entity mention, a score is estimated to predict the likelihood of the candidate concept being the normalized concept.
Note that we use the expanded version of the entity mentions as described above for BM25. 
The following input sequence is fed into BERT for each candidate and mention pair:

\hspace{30pt} [CLS] expanded mention ($m$)[SEP]$c_{i}$[SEP]$syn_{i}$($c_{i}$)[SEP]...$syn_{n}$($c_{i}$)...[SEP]$is$($c_{i}$)[SEP]

Here, $syn_{i}$ refers to any synonym of the candidate concept $c_{i}$ and $is$ refers to the RadLex class to which the candidate concept ($c_{i}$) belongs. 
The order of the synonyms is random.
The main intention behind using the synonyms is that they provide more variation of the candidate concepts and the $is$ provides more information about the candidate concept's class obtained from the `Is-A' attribute in RadLex. 
The final hidden vector corresponding to the [CLS] token in the input sequence is further fed into a single layer network to obtain the estimated probability of how likely the candidate concept is the normalized one.
The probabilities corresponding to all the candidate concepts are then used for ranking. 
Note that the probability score calculated for a particular candidate is independent of the other candidates generated for an entity mention.

\subsubsection{BERT as span detector}
Alternatively, we formulate the normalization problem similar to the BERT framework for a question answering task.
Given an expanded entity mention and its corresponding list of RadLex candidate concepts that is represented as a text sequence, the task is to identify the span of the normalized concept from the candidate list.
The second part in the input sequence (the one followed by the first [SEP]) is constructed by joining all the candidate concept names separated by comma.
The candidate concepts are placed in an arbitrary order to form this sequence.
The final input sequence corresponding to an entity mention and its candidates is represented as follows:

% \phantom{XXXX} [CLS] expanded mention ($m$)[SEP]$c_{1}$, $c_{2}$, ..., $c_{n}$[SEP]
\hspace{30pt} [CLS] expanded mention ($m$)[SEP]$c_{1}$, $c_{2}$, ..., $c_{n}$[SEP]

The scoring mechanism of a candidate span from the sequence of candidate RadLex concepts is same as the implementation in the original BERT paper \cite{devlin2019BERTPretrainingDeep}.
The highest scoring span is identified as the normalized concept for a given mention.

\section*{Experimental Settings and Evaluation}
For both BERT-based normalization methods (re-ranker and span detector), we use the BERT\textsubscript{LARGE} model by initializing the model parameters obtained after pre-training BERT on MIMIC-III clinical notes for $ 320,000 $ steps \cite{si2019EnhancingClinicalConcept}. We fine-tune BERT\textsubscript{LARGE} on our annotated dataset for normalizing the radiological entities. The number of epochs for fine-tuning is decided based on the accuracy of the models on the validation sets. The number of epochs is chosen as 4 for both the normalization models. We use a batch size of 8 for the BERT-based re-ranker model while the batch size of the span detector model is set at 10. We use the cased version of the models. We fine-tune the re-ranker model with a learning rate of 1e-6 and the span detector model with a rate of 3e-5. For the span detector model, we use a maximum sequence length of 384 and a maximum mention length of 64. 

We also utilize BERT\textsubscript{BASE} and BERT\textsubscript{LARGE} models, both pre-trained from clinical notes as mentioned above for automatically detecting the entity spans from the report text. We use the cased version of the models, fine-tune the sequence labeling task for 4 epochs with a learning rate of 2e-5 and maximum sequence length of 128. The batch size used for BERT\textsubscript{BASE} is 24 and BERT\textsubscript{LARGE} is 8.

We evaluate our proposed normalization methods - BERT-based re-ranker and the BERT-based span detector by performing 10-fold cross validation (CV) on our annotated radiology normalization corpus. We create the folds by splitting the corpus at the report level such that the training, validation, and test splits are divided in the ratio of 80-10-10\% respectively. For comparison, we evaluate the predictions of BM25 by averaging the results obtained on the same test folds used for the BERT-based models. Since the focus of this study is on normalizing the various radiological entities to RadLex concepts and not on joint prediction of entity mention spans and their normalized concepts, we conduct all our normalization experiments considering the gold entity mentions. In our experiments, any RID-less (RadLex ID-less) entity mention is represented using a special token-`XXXXX'.

We report the average accuracy of the models using the same fold settings. For BM25 and the BERT-based re-ranker, an exact match between the first ranked concept and the gold annotated concept for an entity mention is considered as a correct prediction. In order to handle cases where no candidates are retrieved by BM25 for a given entity mention, we adjust the performance metric (accuracy) by considering only those as correct predictions when their corresponding gold annotated normalized concept is tagged as `unlinkable' or `RID-less'.

For the BERT span detector model, we take into account an exact match between the predicted span and the gold annotated RadLex concept in the test sets to qualify a prediction as correct. Note that this model can predict any span from the text representing the sequence of comma-separated candidate concepts. Taking this into account, we evaluate the performance of this model in three ways. First, we evaluate using the original predicted text span. In this version, if more than one candidate concept is captured in the predicted span, the prediction is considered incorrect. Second, we employ post-processing of the predicted spans such that if a span contains more than one concept (indicated by comma), we perform an exact match only between the first concept (concept appearing to the left of the first comma in the predicted span) and the gold normalized concept. Third, we conduct a similar evaluation considering the last concept in the predicted span. We further report the average F1-measure of the 10-fold CV on our annotated dataset for detecting the boundaries of the entity mentions given a report sentence to the entity span detection model.

\vspace{-0.25in}

\section*{Results}
The average performance measures of the BERT-based entity span detection system used as a sequence labeler (described in Section \ref{entity_span_detection}) over 10-fold CV are shown in Table \ref{table:bert_span_results}. We notice that the average F1 is increased by around $6.8$ points when the BERT\textsubscript{LARGE} model is used.

We first report the recall of BM25 for candidate generation.
Recall here refers to the percentage of entity mentions for which the list of candidate concepts contains the gold normalized RadLex concept. The recall of BM25 as well as its accuracy in predicting the normalized concepts for 10 and 25 candidates is shown in Table \ref{table:bm25_results}. The average accuracies of 10-fold CV for the BERT-based methods in predicting the correct normalized RadLex concept when provided the 10 candidates generated by BM25 is shown in Table \ref{table:bert_results}. We note that the normalization performance is the highest (accuracy of $77.72$\%) for the BERT-based span detector model when compared to both BM25 and BERT-based re-ranking models. The performance is further improved by $0.7$\% when either the first or the last concept in the predicted span (as predicted by the BERT span detector from the sequence of 10 candidate concepts) is considered as the normalized concept.

\begin{table}[t]
\caption{10-fold CV results for detecting the spans of entity mentions. Both BERT\textsubscript{BASE} and BERT\textsubscript{LARGE} models are pre-trained on MIMIC-III clinical notes.} 
\vspace{-0.25in}
\begin{center}
\begin{tabular}{l|c|c|c}
      
      \hline
      \textbf{Model}&\textbf{Precision(\%)}&\textbf{Recall (\%)}&\textbf{F1} \\
      \hline
      BERT\textsubscript{BASE} & $65.27$ & $73.64$ &  $69.14$ \\
      \hline
      BERT\textsubscript{LARGE} & $72.72$ & $79.64$ & $75.93$ \\
      \hline
\end{tabular}
\end{center}
\label{table:bert_span_results}
\end{table}

\vspace{0.2in}
\begin{table}[t]
\caption{BM25 results in predicting the normalized concepts using 10 and 25 candidate concepts.}
\vspace{-0.25in}
\begin{center}
\begin{tabular}{l|c|c}
      \hline
      \textbf{Metric}&\textbf{10 candidates}&\textbf{25 candidates}\\
      \hline
      Recall (\%) & $88.44$ & $89.72$ \\
 \hline
      Accuracy (\%) &$76.10$ & $76.10$ \\
 \hline
\end{tabular}
\end{center}
\label{table:bm25_results}
\end{table}

\begin{table}[!t]
\caption{10-fold CV results of the proposed BERT-based methods using 10 candidate concepts retrieved by BM25.}
\vspace{-0.25in}
\begin{center}
\begin{tabular}{l|c}
      \hline
      \textbf{Method}&\textbf{Average accuracy (\%)} \\
      \hline
      BERT re-ranker  & $76.50$ \\
 \hline
      BERT span detector (using original predictions) & $77.72$ \\
 \hline
      BERT span detector (first concept in the predicted span as the normalized concept) & $78.43$ \\
 \hline
      BERT span detector (last concept in the predicted span as the normalized concept) & $78.44$ \\
 \hline
\end{tabular}
\end{center}
\label{table:bert_results}
\end{table}

\vspace{-0.25in}

\section*{Discussion}
We create a manually annotated corpus covering a broad range of radiology entity types that are usually of interest for information extraction research. To our knowledge, this is the first study in developing a corpus targeted toward radiology entity normalization. We propose methods for normalizing these entities to an existing lexicon-RadLex.

We also examine the performance of a sequence labeler, based on BERT, for identifying the spans of the entity mentions from the reports. The performance of the span detection system is decent (average F1-score of $75.93$). The moderate performance may be attributed to the incorrect predictions for detecting the composite entities (e.g., ``\textit{right upper lobe}''). Note that the focus of this work is radiology entity normalization, hence we aim to further improve the performance of the entity span detection and develop joint learning methods for predicting the entity spans as well as mapping them to RadLex concepts simultaneously.

Most of the annotation-related challenges are related to the requirement of domain knowledge. For example - ``\textit{Lower pole of the right kidney}'' usually refers to ``\textit{Inferior pole of right kidney}'' and ``\textit{Temporal horns}'' denotes ``\textit{temporal horn of lateral ventricle}''. Besides being a time-consuming process, another generic challenge related to constructing a normalization corpus is the ambiguity involved in annotating composite entity mentions such as ``\textit{right lung apex}''. Another difficulty in the annotation involves dealing with the inconsistencies in the RadLex lexicon. For example, the expression ``\textit{upper lobe of right lung}'' is present in RadLex whereas ``\textit{middle lobe of right lung}'' is not although both are anatomical expressions at the same hierarchical level. The closest term available in RadLex for the middle lobe is ``\textit{middle lobe of lung}''. Also, there are cases where certain entity mentions when expressed using more general terms such as ``\textit{sulci}'' do not appear in RadLex, although their specific types such as ``\textit{hypothalamic sulcus}'' and ``\textit{cardiophrenic sulcus}'' are present in RadLex. 

Our proposed normalization methods achieve satisfactory performance with the highest average accuracy of $78.44$\%. However, we aim to further evaluate the performance of the proposed methods by augmenting the annotated corpus in the future. A brief analysis of the model outputs suggests that the BERT-based models make correct predictions for uncertainty or hedging-related entity mentions such as ``\textit{could indicate}'' that are incorrectly predicted by BM25. Moreover, the BERT span detector model performs better in predicting the normalized concepts for plural entity mentions compared to BERT re-ranker. For instance, BERT span detector predicts ``\textit{lungs}'' as the normalized concept for the mention-``\textit{lungs}'', whereas BERT re-ranker model predicts ``\textit{lung}'' as the mapped concept.
One of the reasons for the moderate performance improvement of the BERT-based models over BM25 may be that our annotated corpus mostly contains different variations of radiological terms unlike social media posts where there are more variations of natural language expressions.

We also intend to conduct an exhaustive set of ablation experiments mainly for the BERT-based re-ranker model utilizing various combination of RadLex knowledge in our later work. Moreover, we plan to use additional techniques for expanding the radiological entity mentions, particularly the ones related to clinical findings (e.g., ``\textit{Scrotal herniation of bowel}''), by leveraging the co-occurring entity information from sources like Wikipedia and medical abstracts.

Among the $151$ entity mentions for which a suitable RadLex concept is not found, some of the most common clinical finding and imaging observation-related entities include - ``\textit{respiratory distress}'', ``\textit{hyaline membrane disease}'', ``\textit{bowel gas pattern}'', ``\textit{cabg}'', ``\textit{portal venous gas}'', ``\textit{mucosal thickening}'', ``\textit{hydropneumothorax}'', ``\textit{ventricular prominence}'', ``\textit{v-fib arrest}'', ``\textit{urinary incontinence}'', ``\textit{reexpansion}'', ``\textit{pleural margins}'', ``\textit{neonatal pneumonia}'', ``\textit{lyme disease}'', ``\textit{intubation}'', ``\textit{guaiac positive stools}'', ``\textit{gonadal shielding}'', ``\textit{fetal lung liquid}'', ``\textit{dyspnea}'', ``\textit{claustrophobia}'', ``\textit{cardiomegaly}'', ``\textit{anxiety}'', ``\textit{altered mental status}'', ``\textit{afib}'', and ``\textit{aeration}''. This can serve as a potential list of terms to expand RadLex in the future.

\section*{Conclusion}
This paper constructs an annotated corpus for radiology entity normalization. The entities cover a variety of important radiological entities including findings, medical devices, and procedures and are mapped to publicly available radiology-specific lexicon-RadLex. This study attempts to standardize the entities commonly extracted by information extraction systems from the radiology reports for various clinical applications. This work also proposes two deep learning-based NLP methods to automatically normalize the entity mentions from the report text. For this, a set of candidate concepts are first retrieved using the BM25 method which are then used by the deep learning methods. Specifically, we configure BERT for the normalization task as a re-ranking model as well as a span detection model by providing the BM25 candidate list as input. We obtain satisfactory results by fine-tuning the BERT-based models on our annotated dataset with the span detector model achieving an accuracy of $78.44$\% in cross validation.

\textbf{Acknowledgments} This work was supported in part by the National Institute of Biomedical Imaging and Bioengineering (R21EB029575), the U.S. National Library of Medicine (R00LM012104), the Patient-Centered Outcomes Research Institute (ME-2018C1-10963), and the Cancer Prevention Research Institute of Texas (RP160015).

\bibliographystyle{vancouver}
\setlength{\bibsep}{0.4pt}
\setlength\itemsep{0.9pt}

\bibliography{amia2020}

\end{document}